\documentclass[sigconf,screen]{acmart}

\AtBeginDocument{%
  }

\setcopyright{cc}
\setcctype{by}
\acmDOI{10.1145/3776734.3794453}
\acmYear{2026}
\copyrightyear{2026}
\acmISBN{979-8-4007-2321-6/2026/03}
\acmConference[HRI Companion '26]{Companion Proceedings of the 21st ACM/IEEE International Conference on Human-Robot Interaction}{March 16--19, 2026}{Edinburgh, Scotland, UK}
\acmBooktitle{Companion Proceedings of the 21st ACM/IEEE International Conference on Human-Robot Interaction (HRI Companion '26), March 16--19, 2026, Edinburgh, Scotland, UK}
\received{2025-12-08}
\received[accepted]{2026-01-12}

\makeatletter
\gdef\@copyrightpermission{
  \begin{minipage}{0.3\columnwidth}
   \href{https://creativecommons.org/licenses/by/4.0/}{\includegraphics[width=0.90\textwidth]{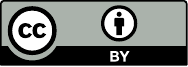}}
  \end{minipage}\hfill
  \begin{minipage}{0.7\columnwidth}
   \href{https://creativecommons.org/licenses/by/4.0/}{This work is licensed under a Creative Commons Attribution International 4.0 License.}
  \end{minipage}
  \vspace{5pt}
}
\makeatother

\begin{document}

\title{Designing Social Robots with Ethical, User-Adaptive Explainability in the Era of Foundation Models}

\author{Fethiye Irmak Doğan}
\orcid{0000-0002-1733-7019}
\authornote{Both authors contributed equally to this research.}
\affiliation{%
  \institution{University of Cambridge}
  \city{Cambridge}
  \country{United Kingdom}
}
\email{fid21@cam.ac.uk}

\author{Alva Markelius}
\orcid{0009-0003-4580-9997}
\authornotemark[1]
\affiliation{%
  \institution{University of Cambridge}
  \city{Cambridge}
  \country{United Kingdom}
}
\email{ajkm4@cam.ac.uk}

\author{Hatice Gunes}
\orcid{0000-0003-2407-3012}
\affiliation{%
  \institution{University of Cambridge}
  \city{Cambridge}
  \country{United Kingdom}
}
\email{hg410@cam.ac.uk}

\renewcommand{\shortauthors}{Dogan, Markelius, and Gunes}
%\end{comment}

\begin{abstract}
Foundation models are increasingly embedded in social robots, mediating not only what they say and do but also how they adapt to users over time. This shift renders traditional ``one-size-fits-all'' explanation strategies especially problematic: generic justifications are now wrapped around behaviour produced by models trained on vast, heterogeneous, and opaque datasets. We argue that ethical, user-adapted explainability must be treated as a core design objective for foundation-model-driven social robotics. We first identify open challenges around explainability and ethical concerns that arise when both adaptation and explanation are delegated to foundation models. Building on this analysis, we propose four recommendations for moving towards user-adapted, modality-aware, and co-designed explanation strategies grounded in smaller, fairer datasets. An illustrative use case of an LLM-driven socially assistive robot demonstrates how these recommendations might be instantiated in a sensitive, real-world domain.
\end{abstract}

\vspace{-0.3em}


\vspace{-0.3em}
\vspace{-0.3em}

\maketitle

\vspace{-0.3em}
\section{Introduction}
\vspace{-0.3em}
Foundation models, particularly large language models (LLMs) and vision–language models (VLMs), are rapidly transforming social robotics. Whereas earlier HRI systems relied on rule-based pipelines, narrowly trained models, or Wizard-of-Oz paradigms, modern systems increasingly use foundation models to generate open-ended language, reason over context, and adapt behaviour in real time, enabling flexible interactions across wellbeing~\cite{10.1145/3712265, 11217833}, education~\cite{Voultsiou_Vrochidou_Moussiades_Papakostas_2025}, assistive~\cite{Hanschmann_Gnewuch_Maedche_2024, axel-skantze}, and companionship domains~\cite{Irfan_Kuoppamäki_Hosseini_Skantze_2025, Irfan_Kuoppamäki_Skantze_2024}. This shift changes how adaptation and explainability are designed: behaviour is now mediated by models trained on vast, heterogeneous, and often opaque corpora, making adaptation alone insufficient. Explainability has long been recognised as a way to render agents’ internal processes accessible to users~\cite{sakai2022explainable}, supporting transparency and understanding~\cite{10.3389/frai.2022.866920}, trust calibration~\cite{siau2018building, edmonds2019tale}, and collaboration~\cite{sridharan2019towards, setchi2020explainable, 9889368, 10.3389/frobt.2022.937772, 9900586}, where explainability is often treated as user-facing justification for a robot’s behaviour and transparency as disclosure of capabilities and limits; yet most work in explainable AI/HRI still assumes relatively static systems and generic users, offering one-size-fits-all templates, visualisations, or rationales that neglect differences in abilities, preferences, cultural backgrounds, or social/emotional/cognitive states. The emergence of foundation models makes this gap more acute: although they can generate fluent and persuasive explanations, these systems often default to generic, surface-level narratives that neither reflect end user diversity nor acknowledge the their limitations and biases.

In this paper, we argue that ethical, user-adapted explainability should be a core design principle %must become a first-class design objective 
for foundation model-driven social robotics, and we outline the ethical risks that arise when both adaptation and explanation are mediated by foundation models. Building on this analysis, we propose four recommendations for moving beyond one-size-fits-all explanations, and to ground these ideas, we illustrate how our recommendations could be instantiated in the concrete use case of an LLM-driven socially assistive robot, where explanation, adaptation, and vulnerability may intersect. Finally, we discuss key considerations and concerns for adaptive explanation generation. Our aim is not to offer definitive technical solutions, but to map an emerging problem space and set an agenda for future research on ethical, explainable adaptation in foundation model-driven social robotics.

\vspace{-0.3em}
\section{Related Work}
\vspace{-0.3em}

\textbf{Foundation Models for Adaptive Social Robotics:}
Traditional HRI relied on rigid rule-based systems, narrowly trained supervised models, or Wizard-of-Oz setups, which enabled basic functionality but suffered from poor scalability, poor generalisation across contexts, and limited naturalistic adaptation \cite{leusmann2024comparing, bejarano2024hardships}. Foundation models, particularly LLMs, have begun to transform social robotics by enabling more naturalistic language, contextual reasoning, and behavioural adaptation \cite{Shi_Landrum_O’Connell_Kian_Pinto-Alva_Shrestha_Zhu_Matarić_2024}. Empirical deployments already illustrate this shift toward adaptive interaction. Socially assistive robots use foundation models for personalised well-being coaching \cite{10.1145/3712265}, special-education support \cite{Voultsiou_Vrochidou_Moussiades_Papakostas_2025}, motor-disability assistance via voice control \cite{Padmanabha_Yuan_Gupta_Karachiwalla_Majidi_Admoni_Erickson_2024}, cognitive interventions for autistic users \cite{Bertacchini_Demarco_Scuro_Pantano_Bilotta_2023}, and companionship for older adults \cite{Irfan_Kuoppamäki_Hosseini_Skantze_2025, Irfan_Kuoppamäki_Skantze_2024}. Commercial and public-facing robots use LLMs as retail assistants \cite{Hanschmann_Gnewuch_Maedche_2024}, museum guides \cite{axel-skantze}, and real-time language translators \cite{Jokinen_Wilcock_2024}. In multi-party settings, LLMs manage socially appropriate turn-taking and group dynamics \cite{Addlesee_Cherakara_Nelson_2024, 10.1145/3544549.3585602}. Personalisation extends to tailored speech patterns \cite{onorati2023creating, sevilla2023using}, empathetic responses, and contextually generated non-verbal behaviours, including emotion expression \cite{Mishra_Verdonschot_Hagoort_Skantze_2023} and emoji-triggered physical actions \cite{Wang_Reisert_Nichols_Gomez_2024}. Multimodal integration pipelines now translate gaze, facial expressions, posture, and scene objects into natural-language prompts, allowing LLMs to produce highly context-sensitive robot responses \cite{Wang_Hasler_Tanneberg_Ocker_Joublin_Ceravola_Deigmoeller_Gienger_2024}. However, despite their potential, foundation models are not sufficient to capture all nuances of social interactions. For instance, LLM-based decisions for socially appropriate robot actions are found to be insufficient and should be improved with user explanations to tailor them to users' individual expectations and preferences~\cite{11127826}. %LLM-based decisions for socially appropriate robot actions, while markedly more adaptive than earlier paradigms, remain inherently opaque and normatively pre-committed. 
These findings highlight that without mechanisms that render these decisions legible and contestable to users, adaptation risks reproducing misalignments and exclusions.

\textbf{Adaptive Explanations for Social Robotics:}
Explainability has been proposed as a key mechanism for rendering agents’ and robots’ internal decision processes accessible and understandable to people~\cite{sakai2022explainable}, thereby reducing the communication gap and enabling mutual understanding between users and robots~\cite{10.1145/3610978.3638154}. In the context of social robotics, explanations are expected to be correct, concise, and unambiguous, and to be presented in a manner that is sensitive to the interaction context~\cite{ALI2023101805}. Prior work has shown that explainability in HRI and social robotics is closely tied to increased perceived trust in robots~\cite{siau2018building, edmonds2019tale}, greater accountability of autonomous systems~\cite{barredoarrieta2020ExplainableArtificialIntelligence}, and improved efficiency in human–robot interaction and collaboration~\cite{sridharan2019towards, setchi2020explainable, 9889368, 10.3389/frobt.2022.937772, 9900586}. Explanations in HRI should be designed for end users to support understanding of a robot’s cognitive and decision-making processes. Despite their importance, research on adapting explanations to task and user characteristics is limited. Guerdan et al.~\cite{10.5555/3398761.3398890, 9607734} found that explanations add little value for simple or transparent tasks but improve performance and experience in complex interactions, also accounting for users’ affective responses. Similarly, Yadollahi et al.~\cite{yadollahi2025expectations} highlight the importance of accounting for users' expectations when designing explanation strategies in HRI. Indeed, these studies represent early but important steps towards adapting robot explanations to task demands and user characteristics. However, there is still a need for systematic investigations of fine-grained adaptation strategies, for example, across different demographic groups, backgrounds, and age ranges, as well as with respect to specific interaction scenarios and environmental conditions. %Indeed, the need for user-sensitive explanations intersects with ethical concerns, such as those accounted for in the next section.
This need is also closely tied to ethical concerns as discussed in the next section.

\textbf{Ethical Considerations for Adaptive Social Robotics:} Based on recent scholarship on foundation-model-driven HRI \cite{Markelius, Shi_Landrum_O’Connell_Kian_Pinto-Alva_Shrestha_Zhu_Matarić_2024, Williams_Matuszek_Mead_Depalma_2024, Atuhurra_2024} we identify a range of ethical considerations. Firstly, safety concerns arise from the opacity and unpredictability of model outputs in real-time interaction \cite{leusmann2024comparing}, with LLM-driven robots vulnerable to adversarial prompting and other reliability failures \cite{Abbo_Desideri_Belpaeme_Spitale_2025}. Normativity concerns stem from the tendency of foundation models to reproduce hegemonic worldviews and flatten pluralistic knowledge systems, producing inconsistencies and normative bias in subjective or value-laden interactions \cite{Sherer_McPherson_Mohanty_Santé_Gandolfi_Romeo_Suglia_2025, Williams_Matuszek_Mead_Depalma_2024}. Furthermore, hyper-personalisation introduces risks of emotional influence and subtle manipulation, as ingratiating model behaviours reinforce user trust, shape intentions, and commodify affective exchanges \cite{Chaudhary_Penn_2024, Carro_2024}. Finally, persistent bias in model outputs, ranging from racialised stereotypes \cite{Abid_Farooqi_Zou_2021} and geopolitical distortions \cite{Whatpale} to anti-LGBTQ+ bias \cite{Felkner_Chang_Jang_May_2024} and gendered rationalisations \cite{Kotek_Dockum_Sun_2023} further reflects structural epistemic centralisation, with harmful consequences for marginalised groups \cite{Atari_Xue_Park_Blasi_Henrich_2023}. Ostrowski et al.~\cite{ostrowski2022ethics}, using Design Justice principles~\cite{costanza2020design}, show that ethics and equity remain marginal in HRI, highlighting risks such as exclusionary design, beneficiary misalignment, bias amplification, power asymmetries, dependence, deception, and regulatory gaps—issues highly relevant for foundation-model-driven adaptation. Complementing this, Markelius~\cite{Markelius} maps ethical hazards at the LLM–social-robotics intersection, noting that physical embodiment amplifies classic LLM risks (misinformation, bias, emotional disruption) and introduces new concerns around anthropomorphism, relational asymmetry, and commodification of intimacy. 
Indeed, a central concern is how adaptive, foundation-model-driven robots may disproportionately harm marginalised populations~\cite{Markelius} and how intersectional socio-technical factors shape both system behaviour and the distribution of harms~\cite{paterson2024robot, seaborn2022not, winkle2023feminist}.

\vspace{-0.3em}
\section{Open Challenges}
\vspace{-0.3em}

\begin{figure}
    \centering
    \includegraphics[width=0.95\linewidth]{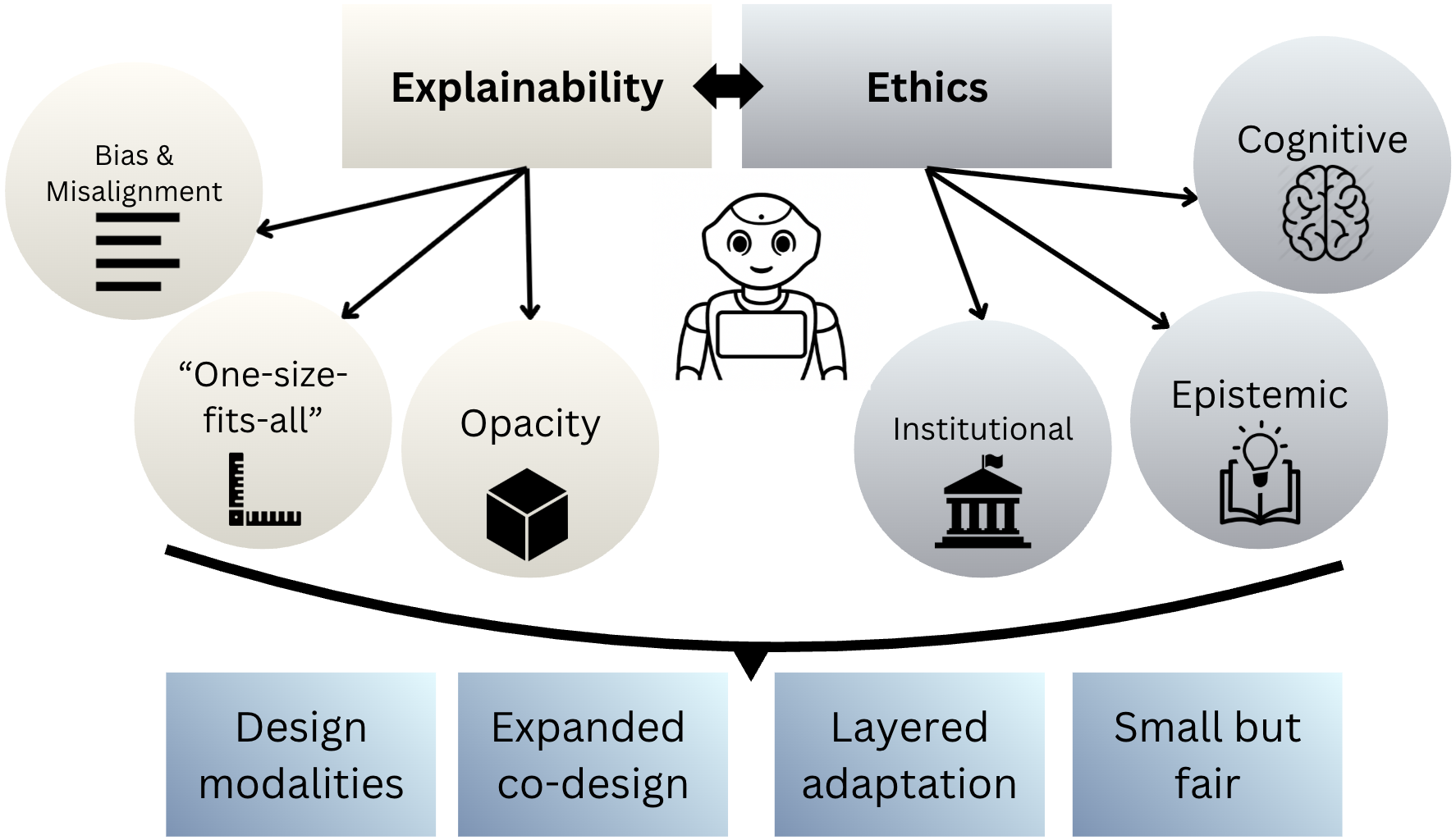}
    \vspace{-0.8em}
    \caption{Open challenges in ethical and explainable adaptation in foundation model-driven social robotics and our proposed avenues of realisation and recommendations.}
    \Description{The diagram of open challenges and recommendations for ethical and user-adaptive explainability for foundation model-driven social robots. At the top, two adjoining rectangles labelled ``Explainability'' and ``Ethics'' are connected by a double-headed arrow. There is a cartoon social robot in the centre. On the left side (explainability), arrows point to three circles representing open challenges: ``Bias \& Misalignment'', ``One-size fits-all'', and ``Opacity''. On the right side (ethics), arrows point to three circles representing ethical challenges: ``Cognitive'', ``Institutional'', and ``Epistemic''. Along the bottom, a curved line points down to four rectangles showing our proposed recommendations: ``Design modalities'', ``Expanded co-design'', ``Layered adaptation'', and ``Small but fair''.}
    \label{fig}
\vspace{-1em}
\end{figure}

This section outlines the key challenges identified for explainability and ethics in foundation-model-driven social robotics -- see Figure~\ref{fig}.

\vspace{-0.7em}
\subsection{User-Adaptive Explainability}% for Foundation Model-driven Social Robotics}
\vspace{-0.3em}

Prior to the emergence of LLMs and foundation models, adaptation in social robotics was typically grounded in data collected from small-scale, controlled user studies~\cite{martins2019user}. These systems were often designed and tuned using interaction data from relatively homogeneous participant groups in laboratory or field settings, and adaptation mechanisms were explicitly crafted to generalise from this limited, well-characterised population to similar users~\cite{seaborn2023not}. In such setups, both the scope of adaptation and the corresponding explanations could be carefully constrained: designers knew the characteristics of the training population, the task domain was narrow, and the space of possible behaviours and explanations was relatively bounded. As a result, user-adapted explanations are tightly coupled to specific scenarios and to user models that approximate only a small subset of the broader population.

In contrast, foundation model-driven social robots disrupt many of these assumptions and introduce new open challenges for user-adapted explainability. 
They are typically powered by models trained on internet-scale data, which radically expand the space of possible behaviours and explanations while reducing designers’ control over training signals and internal representations, and introducing ethical hazards such as normative cognitive and epistemic values and liability difficulties. 
We highlight three interrelated challenges:

\textbf{(1) Pre-training biases and misalignment.} The internet-scale corpora used to train foundation models encode \textbf{implicit assumptions, biases, and interaction patterns induced by their pre-training data}, which may not align with the needs, preferences, or expectations of individual users in situated HRI contexts. %While such models enable highly generalisable and open-ended interaction capabilities, these inherited patterns make genuine user-specific adaptation and, crucially, explaining such adaptation substantially more challenging.
While such models enable highly generalisable, open-ended interaction, these inherited patterns make genuine user-specific adaptation—and, crucially, explaining it—substantially more challenging.

\textbf{(2) Persistence of one-size-fits-all explanations.} Current systems often default to generic, \textbf{one-size-fits-all explanations} that fail to reflect the user’s background, abilities, or mental model of the robot. This limits the potential of foundation model-driven social robots to provide explanations that are genuinely supportive, accessible, and meaningful for diverse user groups.

\textbf{(3) Opacity of adaptive mechanisms.} How to systematically personalise explanations for diverse end users, while remaining faithful to the underlying foundation model’s behaviour and maintaining transparency about what is being adapted and why, remains an open and largely unexplored research gap. This includes understanding how to calibrate explanations for different demographics, cognitive and cultural backgrounds, and interaction contexts, and how to expose the logic of adaptation itself in ways that are comprehensible, trustworthy, and ethically aligned.

\vspace{-0.8em}
\subsection{Ethical Concerns}% for Foundation Model-driven Social Robotics}
\vspace{-0.3em}

In this section, we take a preliminary step toward mapping the ethical implications that arise when user adaptation and explanation in social robotics are mediated through foundation models. Unlike earlier adaptive HRI systems trained on small, homogeneous datasets, large-scale pre-trained models introduce a qualitatively different ethical complexity, as they encode normative assumptions, epistemic exclusions, and value hierarchies that may not align with users’ situated needs \cite{palacios2025intersection, hila2025epistemological, Atari_Xue_Park_Blasi_Henrich_2023, Markelius, Sherer_McPherson_Mohanty_Santé_Gandolfi_Romeo_Suglia_2025}. Building on the explainability challenges outlined above, we examine the ethical risks associated with foundation-model-driven adaptation and explanation, focusing on misalignment, systemic bias, and the uneven distribution of harm. We analyse these issues across three perspectives.% cognitive, epistemic, and institutional.

Firstly, from a \textbf{cognitive standpoint}, adaptation in foundation-model-driven social robotics risks causing misalignment between the robot’s inferred user model and the user’s actual mental states, intentions, and emotions. Foundation models generate user representations through opaque probabilistic reconstructions of human behaviour that routinely conflate correlation with moral or psychological validity, projecting culturally dominant cognitive styles onto diverse users~\cite{Atari_Xue_Park_Blasi_Henrich_2023}. One prominent example concerns neurodivergence and disability~\cite{markscoping}. Trained primarily on neurotypical behavioural patterns, foundation models may interpret neurodivergent gaze, prosody, or language as deficits, or pathalogise user~\cite{dehnert2024ability, Williams_2021}, while their explanations can further reinforce these misinterpretations by framing neurodivergent behaviour as a sign of discomfort (e.g., ``I am giving you space because you seemed uncomfortable''), making the bias more dominant yet less visible. As a result, a robot might adapt inappropriately, instantiating cognitive ableism under the guise of personalised adaptation~\cite{Frennert_Persson_Skavron_2024, 10.1145/3613904.3642798}.

Secondly, \textbf{epistemically}, the deployment of foundation models affects a covert centralisation of knowledge authority that undermines the situated, co-constructed nature of human–robot knowledge practices. By privileging patterns extracted from historically contingent and power-laden datasets, these models enact systemic epistemic exclusion \cite{hila2025epistemological}, marginalising non-hegemonic ways of knowing, reasoning, and expressing agency. Explanations generated post-hoc to justify adaptive behaviours thus function less as genuine accountability mechanisms than as performative rationalisations that obscure the contested normative commitments embedded in the model \cite{palacios2025intersection}. Users, particularly those from underrepresented linguistic \cite{khanna2025invisible}, Indigenous \cite{hanson2025indigenous}, or neurodivergent communities are positioned as epistemic subjects whose validity is measured against an unacknowledged universalist benchmark, foreclosing the possibility of mutual epistemic adjustment and reinforcing existing hierarchies of credibility.

Finally, \textbf{institutionally}, the integration of foundation models into social robotics accelerates a commodification and externalisation of ethical responsibility \cite{kunz2020}. Developers can appeal to the so-called neutrality and scale of pre-training as a moral alibi, diffusing accountability across global data supply chains while concentrating control in the hands of a small number of technology providers \cite{brown2023allocating, cobbe2023understanding}. This creates asymmetric power relations: harms arising from biased adaptation and misleading or partial explanations are disproportionately borne by marginalised groups and are framed as inevitable by-products of ``general-purpose'' systems rather than as foreseeable consequences of design choices \cite{Markelius, Sherer_McPherson_Mohanty_Santé_Gandolfi_Romeo_Suglia_2025}. The resulting regulatory vacuum permits the quiet delegation of normative governance to private entities whose incentive structures may prioritise, i.e., engagement metrics over distributional justice, thereby normalising a corporate control in which the explanations users receive are shaped by commercial interests, pre-empting democratic contestation of the values inscribed in everyday HRI \cite{ostrowski2022ethics, rakova2023terms}.

User-adapted explanations can help address these risks by supporting alignment, contestability, and accountability in interaction; however, these ethical issues are complex and structural, and cannot be resolved through explanation design alone.

\vspace{-0.8em}
\section{Towards Realisation and Recommendations}
\vspace{-0.3em}

At the core of our argument is that ``one-size-fits-all'' explanations are incompatible with the diverse users and contexts of foundation model-driven social robotics. 
Explanations should be tailored to users and tasks, as channels—text, visual, and nonverbal—may be differently preferred and vary across applications and interactions. 
This also addresses the previously outlined ethical considerations and supports cognitive and epistemic diversity. We outline four recommendations for user-adapted explainability.

\textbf{Recommendation 1: Design modality- and user-aware explanation strategies.}  
Robot explanations should be tailored to users’ abilities and preferences across modalities. Complex language may hinder understanding for children or people with developmental language disorders, who may benefit from simplified text or visual supports. Conversely, visual explanations may be inaccessible to users with visual impairments. Nonverbal cues can be effective in some cultural contexts but ambiguous or inappropriate in others. Systems should adaptively select and combine channels based on user, task, and sociocultural factors.

\textbf{Recommendation 2: Employ co-design to elicit explanation needs and adaptation criteria.}  
Co-design approaches have been shown to be central for inclusive social robot design~\cite{markelius2025stakeholder}, and we argue that this should extend to explanation, adaptation mechanisms and ethical considerations. Users should be involved in determining how robots explain their behaviours, what ethical aspects are important, what aspects should be adapted, and how this adaptation should be communicated. Co-design can help surface users’ cognitive, emotional,sensory and epistemic perspectives, and inform explanation and adaptation that are neither overwhelming nor patronising. Beyond traditional design dimensions such as robot appearance or voice, participatory processes can guide the granularity, timing, and modality of explanations and ethical aspects, allowing systems to adapt to diverse epistemic and cognitive profiles, rather than relying on designer assumptions.

\textbf{Recommendation 3: Integrate layered adaptation that balances user preferences and technical feasibility.}  
Not all user preferences can be met with current robotic and AI capabilities, and per-user fine-tuning may be impractical, e.g., latency, on-device compute, and data availability can limit per-user models. We advocate layered adaptation strategies that balance user needs with system constraints. Transfer learning from similar user groups can provide a starting point, while parameter-efficient techniques (e.g., low-rank or adapter-based fine-tuning~\cite{hu2022lora}) enable lightweight adaptation to user clusters, allowing finer personalisation as more data becomes available. These strategies help avoid ethically fraught practices like over-personalisation, unjustified data harvesting, or reinforcing inequities, while still allowing finer personalisation as more interaction data is responsibly obtained. The goal is to adapt as far as is technically and ethically feasible, without over-promising, over-fitting to privileged users, or compromising robustness.

\textbf{Recommendation 4: Prioritise smaller, fairer datasets over indiscriminate scaling.}  
The “big models, big data” paradigm does not automatically support user-adapted explainability. Large, uncurated datasets can amplify biases and produce overgeneralised explanations that overlook minority or vulnerable users. Smaller, curated datasets, as shown in HRI studies~\cite{cheong2024small}, enable fairer adaptation. We recommend combining foundation models with co-designed data collection to foreground underrepresented users, ensuring explanations are both effective and ethically grounded.

\vspace{-0.8em}
\subsection{Illustrative Use Case}%: LLM-Driven Socially Assistive Robot}
\vspace{-0.3em}

To concretise our recommendations, we consider an LLM-driven socially assistive robot that supports users with diverse abilities, access needs, and lived experiences (e.g., disabled users, older adults, LGBTQ+ and Indigenous communities) in schools, care homes, and community or clinical contexts. In these settings, a ``one-size-fits-all'' explanation strategy is clearly inadequate: users differ in their emotional states, prior experiences, and levels of trust in technology.

An LLM-driven socially assistive robot should therefore adapt explanation format and complexity to both the user group and the current interaction state (\textbf{R1}). For example, children or users with developmental language disorder may benefit from short, concrete explanations with visual aids, whereas older adults may prefer text- or speech-based rationales; in some cultural contexts, nonverbal cues (e.g., gaze, gesture) may be especially meaningful. Emotional state is also critical: elaborate explanations may be less effective for users experiencing negative emotions or high cognitive load~\cite{10.1007/978-3-032-08333-3_11}. These design choices should be informed through co-design with users and professionals, who can jointly specify modalities, timings, and levels of detail (\textbf{R2}). In parallel, per-user fine-tuning of explanation strategies may be unrealistic, e.g., if a school-based robot rotates across many classrooms, or a wellbeing companion has short and infrequent check-in conversations with patients. A practical alternative is layered adaptation, where initial models are adapted to broad user clusters (e.g., age group, accessibility needs, cultural background) and then refined as more interaction data and feedback become available (\textbf{R3}). To support this process, explanation strategies can be constrained and evaluated using small, curated datasets co-designed with relevant, particularly underrepresented, communities (\textbf{R4}). In this way, adaptive explanations are anchored in both powerful foundation models and situated, community-informed preferences.

\vspace{-0.8em}
\section{Considerations and Concerns}
\vspace{-0.3em}

Explainability is often promoted as a means to foster appropriate trust; however, explanations can also create a misleading illusion of competence that exceeds a robot’s actual capabilities~\cite{ehsan2024explainability}. This risk is amplified when explanations are adapted to specific user groups: carefully tailored, plausible-sounding narratives may make the system appear more accurate, robust, or context-aware than it truly is. In the case of foundation model-driven social robots, which can generate fluent and context-sensitive language, there is a particular danger that users infer a rich internal understanding where there is only pattern matching~\cite{shojaee2025illusion, 10.1145/3442188.3445922}. In such settings, explanations can be deceptive and foster unwarranted confidence in a system's outputs~\cite{10.5555/3709347.3743971}, and this risk is likely amplified for overly elaborate or personalised explanations which may promote over-trust, obscure system limitations, and hinder critical scrutiny. Ethical adaptation must explicitly account for these risks by constraining explanation content to what the system can reliably justify, clearly signalling uncertainty and competence boundaries, and avoiding design choices that deliberately or inadvertently amplify anthropomorphism.

Another concern is that explanations are not universally beneficial, nor always necessary. Existing work indicates that, for relatively simple tasks or highly transparent interactions, explanations may provide limited added value, whereas they can be more beneficial in complex systems~\cite{10.5555/3398761.3398890, 9607734}. In some contexts, offering explanations may even introduce cognitive overload, distraction, or emotional strain—particularly in sensitive domains such as wellbeing or mental health, which can cause ethical hazards and harm. Accordingly, user-adapted explainability should incorporate the possibility of minimal or on-demand explanations, rather than assuming that more explanation is always better. This entails designing mechanisms that can (i) assess when explanations are likely to be helpful or harmful, (ii) calibrate their depth and frequency to the user’s current needs and state, and (iii) remain transparent about the limits of both the robot’s reasoning and its adaptation process.

\vspace{-0.8em}
\begin{acks}
\vspace{-0.3em}
\textbf{Funding:} FID \& HG were supported in part by CHANSE and NORFACE through the MICRO project, funded by ESRC/UKRI (grant ref. UKRI572). AM is supported by the Cambridge International Trust Scholarship. \textbf{Contributions:} Conceptualisation, Resources, Writing: FID, AM, HG; Visualisation:~AM; Supervision: HG.
\end{acks}

\balance
\bibliographystyle{ACM-Reference-Format}
\bibliography{refs}


\begin{thebibliography}{76}

%%% the Web address.

\ifx \showCODEN    \undefined \def \showCODEN     #1{\unskip}     \fi
\ifx \showISBNx    \undefined \def \showISBNx     #1{\unskip}     \fi
\ifx \showISBNxiii \undefined \def \showISBNxiii  #1{\unskip}     \fi
\ifx \showISSN     \undefined \def \showISSN      #1{\unskip}     \fi
\ifx \showLCCN     \undefined \def \showLCCN      #1{\unskip}     \fi
\ifx \shownote     \undefined \def \shownote      #1{#1}          \fi
\ifx \showarticletitle \undefined \def \showarticletitle #1{#1}   \fi
\ifx \showURL      \undefined \def \showURL       {\relax}        \fi
% invisible to TeX
\providecommand\bibfield[2]{#2}
\providecommand\bibinfo[2]{#2}
\providecommand\natexlab[1]{#1}
\providecommand\showeprint[2][]{arXiv:#2}

\bibitem[Abbasi et~al\mbox{.}(2025)]%
        {11217833}
\bibfield{author}{\bibinfo{person}{Nida~Itrat Abbasi}, \bibinfo{person}{Fethiye~Irmak Dogan}, \bibinfo{person}{Guy Laban}, \bibinfo{person}{Joanna Anderson}, \bibinfo{person}{Tamsin Ford}, \bibinfo{person}{Peter~B. Jones}, {and} \bibinfo{person}{Hatice Gunes}.} \bibinfo{year}{2025}\natexlab{}.
\newblock \showarticletitle{Robot-Led Vision Language Model Wellbeing Assessment of Children}. In \bibinfo{booktitle}{\emph{2025 34th IEEE International Conference on Robot and Human Interactive Communication (RO-MAN)}}. \bibinfo{pages}{59--64}.
\newblock
\href{https://doi.org/10.1109/RO-MAN63969.2025.11217833}{doi:\nolinkurl{10.1109/RO-MAN63969.2025.11217833}}


\bibitem[Abbo et~al\mbox{.}(2025)]%
        {Abbo_Desideri_Belpaeme_Spitale_2025}
\bibfield{author}{\bibinfo{person}{Giulio~Antonio Abbo}, \bibinfo{person}{Gloria Desideri}, \bibinfo{person}{Tony Belpaeme}, {and} \bibinfo{person}{Micol Spitale}.} \bibinfo{year}{2025}\natexlab{}.
\newblock \showarticletitle{“Can you be my mum?”: Manipulating Social Robots in the Large Language Models Era}.
\newblock \bibinfo{journal}{\emph{arXiv Preprint}} \bibinfo{number}{arXiv:2501.04633} (\bibinfo{date}{Jan.} \bibinfo{year}{2025}).
\newblock
\href{https://doi.org/10.48550/arXiv.2501.04633}{doi:\nolinkurl{10.48550/arXiv.2501.04633}}
\newblock
\shownote{arXiv:2501.04633 [cs]}.


\bibitem[Abid et~al\mbox{.}(2021)]%
        {Abid_Farooqi_Zou_2021}
\bibfield{author}{\bibinfo{person}{Abubakar Abid}, \bibinfo{person}{Maheen Farooqi}, {and} \bibinfo{person}{James Zou}.} \bibinfo{year}{2021}\natexlab{}.
\newblock \showarticletitle{Persistent Anti-Muslim Bias in Large Language Models}. In \bibinfo{booktitle}{\emph{Proceedings of the 2021 AAAI/ACM Conference on AI, Ethics, and Society}} \emph{(\bibinfo{series}{AIES ’21})}. \bibinfo{publisher}{Association for Computing Machinery}, \bibinfo{address}{New York, NY, USA}, \bibinfo{pages}{298–306}.
\newblock
\showISBNx{978-1-4503-8473-5}
\href{https://doi.org/10.1145/3461702.3462624}{doi:\nolinkurl{10.1145/3461702.3462624}}


\bibitem[Addlesee et~al\mbox{.}(2024)]%
        {Addlesee_Cherakara_Nelson_2024}
\bibfield{author}{\bibinfo{person}{Angus Addlesee}, \bibinfo{person}{Neeraj Cherakara}, \bibinfo{person}{Nivan Nelson}, \bibinfo{person}{Daniel Hernández~García}, \bibinfo{person}{Nancie Gunson}, \bibinfo{person}{Weronika Sieińska}, \bibinfo{person}{Marta Romeo}, \bibinfo{person}{Christian Dondrup}, {and} \bibinfo{person}{Oliver Lemon}.} \bibinfo{year}{2024}\natexlab{}.
\newblock \showarticletitle{A Multi-party Conversational Social Robot Using LLMs}. In \bibinfo{booktitle}{\emph{Companion of the 2024 ACM/IEEE International Conference on Human-Robot Interaction}} \emph{(\bibinfo{series}{HRI ’24})}. \bibinfo{publisher}{Association for Computing Machinery}, \bibinfo{address}{New York, NY, USA}, \bibinfo{pages}{1273–1275}.
\newblock
\showISBNx{979-8-4007-0323-2}
\href{https://doi.org/10.1145/3610978.3641112}{doi:\nolinkurl{10.1145/3610978.3641112}}


\bibitem[Ali et~al\mbox{.}(2023)]%
        {ALI2023101805}
\bibfield{author}{\bibinfo{person}{Sajid Ali}, \bibinfo{person}{Tamer Abuhmed}, \bibinfo{person}{Shaker El-Sappagh}, \bibinfo{person}{Khan Muhammad}, \bibinfo{person}{Jose~M. Alonso-Moral}, \bibinfo{person}{Roberto Confalonieri}, \bibinfo{person}{Riccardo Guidotti}, \bibinfo{person}{Javier {Del Ser}}, \bibinfo{person}{Natalia Díaz-Rodríguez}, {and} \bibinfo{person}{Francisco Herrera}.} \bibinfo{year}{2023}\natexlab{}.
\newblock \showarticletitle{Explainable Artificial Intelligence (XAI): What we know and what is left to attain Trustworthy Artificial Intelligence}.
\newblock \bibinfo{journal}{\emph{Information Fusion}}  \bibinfo{volume}{99} (\bibinfo{year}{2023}), \bibinfo{pages}{101805}.
\newblock
\showISSN{1566-2535}
\href{https://doi.org/10.1016/j.inffus.2023.101805}{doi:\nolinkurl{10.1016/j.inffus.2023.101805}}


\bibitem[Atari et~al\mbox{.}(2023)]%
        {Atari_Xue_Park_Blasi_Henrich_2023}
\bibfield{author}{\bibinfo{person}{Mohammad Atari}, \bibinfo{person}{Mona Xue}, \bibinfo{person}{Peter Park}, \bibinfo{person}{Damián Blasi}, {and} \bibinfo{person}{Joseph Henrich}.} \bibinfo{year}{2023}\natexlab{}.
\newblock \showarticletitle{Which Humans?}
\newblock  (\bibinfo{date}{Sept.} \bibinfo{year}{2023}).
\newblock
\href{https://doi.org/10.31234/osf.io/5b26t}{doi:\nolinkurl{10.31234/osf.io/5b26t}}


\bibitem[Atuhurra(2024)]%
        {Atuhurra_2024}
\bibfield{author}{\bibinfo{person}{Jesse Atuhurra}.} \bibinfo{year}{2024}\natexlab{}.
\newblock \showarticletitle{Leveraging Large Language Models in Human-Robot Interaction: A Critical Analysis of Potential and Pitfalls}.
\newblock \bibinfo{journal}{\emph{arXiv Preprint}} \bibinfo{number}{arXiv:2405.00693} (\bibinfo{date}{Nov.} \bibinfo{year}{2024}).
\newblock
\href{https://doi.org/10.48550/arXiv.2405.00693}{doi:\nolinkurl{10.48550/arXiv.2405.00693}}
\newblock
\shownote{arXiv:2405.00693 [cs]}.


\bibitem[Axelsson and Skantze(2023)]%
        {axel-skantze}
\bibfield{author}{\bibinfo{person}{Agnes Axelsson} {and} \bibinfo{person}{Gabriel Skantze}.} \bibinfo{year}{2023}\natexlab{}.
\newblock \showarticletitle{Do You Follow? A Fully Automated System for Adaptive Robot Presenters}. In \bibinfo{booktitle}{\emph{Proceedings of the 2023 ACM/IEEE International Conference on Human-Robot Interaction}} (Stockholm, Sweden) \emph{(\bibinfo{series}{HRI '23})}. \bibinfo{publisher}{Association for Computing Machinery}, \bibinfo{address}{New York, NY, USA}, \bibinfo{pages}{102–111}.
\newblock
\showISBNx{9781450399647}
\href{https://doi.org/10.1145/3568162.3576958}{doi:\nolinkurl{10.1145/3568162.3576958}}


\bibitem[Barredo~Arrieta et~al\mbox{.}(2020)]%
        {barredoarrieta2020ExplainableArtificialIntelligence}
\bibfield{author}{\bibinfo{person}{Alejandro Barredo~Arrieta}, \bibinfo{person}{Natalia {D{\'i}az-Rodr{\'i}guez}}, \bibinfo{person}{Javier Del~Ser}, \bibinfo{person}{Adrien Bennetot}, \bibinfo{person}{Siham Tabik}, \bibinfo{person}{Alberto Barbado}, \bibinfo{person}{Salvador Garcia}, \bibinfo{person}{Sergio {Gil-Lopez}}, \bibinfo{person}{Daniel Molina}, \bibinfo{person}{Richard Benjamins}, \bibinfo{person}{Raja Chatila}, {and} \bibinfo{person}{Francisco Herrera}.} \bibinfo{year}{2020}\natexlab{}.
\newblock \showarticletitle{Explainable {{Artificial Intelligence}} ({{XAI}}): {{Concepts}}, Taxonomies, Opportunities and Challenges toward Responsible {{AI}}}.
\newblock \bibinfo{journal}{\emph{Information Fusion}}  \bibinfo{volume}{58} (\bibinfo{date}{June} \bibinfo{year}{2020}), \bibinfo{pages}{82--115}.
\newblock
\showISSN{15662535}
\href{https://doi.org/10.1016/j.inffus.2019.12.012}{doi:\nolinkurl{10.1016/j.inffus.2019.12.012}}


\bibitem[Bejarano et~al\mbox{.}(2024)]%
        {bejarano2024hardships}
\bibfield{author}{\bibinfo{person}{Alexandra Bejarano}, \bibinfo{person}{Saad Elbeleidy}, \bibinfo{person}{Terran Mott}, \bibinfo{person}{Sebastian Negrete-Alamillo}, \bibinfo{person}{Luis~Angel Armenta}, {and} \bibinfo{person}{Tom Williams}.} \bibinfo{year}{2024}\natexlab{}.
\newblock \showarticletitle{Hardships in the Land of Oz: Robot Control Challenges Faced by HRI Researchers and Real-World Teleoperators}. In \bibinfo{booktitle}{\emph{2024 33rd IEEE International Conference on Robot and Human Interactive Communication (ROMAN)}}. IEEE, \bibinfo{pages}{1914--1921}.
\newblock


\bibitem[Bender et~al\mbox{.}(2021)]%
        {10.1145/3442188.3445922}
\bibfield{author}{\bibinfo{person}{Emily~M. Bender}, \bibinfo{person}{Timnit Gebru}, \bibinfo{person}{Angelina McMillan-Major}, {and} \bibinfo{person}{Shmargaret Shmitchell}.} \bibinfo{year}{2021}\natexlab{}.
\newblock \showarticletitle{On the Dangers of Stochastic Parrots: Can Language Models Be Too Big?}. In \bibinfo{booktitle}{\emph{Proceedings of the 2021 ACM Conference on Fairness, Accountability, and Transparency}} (Virtual Event, Canada) \emph{(\bibinfo{series}{FAccT '21})}. \bibinfo{publisher}{Association for Computing Machinery}, \bibinfo{address}{New York, NY, USA}, \bibinfo{pages}{610–623}.
\newblock
\showISBNx{9781450383097}
\href{https://doi.org/10.1145/3442188.3445922}{doi:\nolinkurl{10.1145/3442188.3445922}}


\bibitem[Bertacchini et~al\mbox{.}(2023)]%
        {Bertacchini_Demarco_Scuro_Pantano_Bilotta_2023}
\bibfield{author}{\bibinfo{person}{Francesca Bertacchini}, \bibinfo{person}{Francesco Demarco}, \bibinfo{person}{Carmelo Scuro}, \bibinfo{person}{Pietro Pantano}, {and} \bibinfo{person}{Eleonora Bilotta}.} \bibinfo{year}{2023}\natexlab{}.
\newblock \showarticletitle{A social robot connected with chatGPT to improve cognitive functioning in ASD subjects}.
\newblock \bibinfo{journal}{\emph{Frontiers in Psychology}}  \bibinfo{volume}{14} (\bibinfo{year}{2023}), \bibinfo{pages}{01--22}.
\newblock
\showISSN{1664-1078}
\href{https://doi.org/10.3389/fpsyg.2023.1232177}{doi:\nolinkurl{10.3389/fpsyg.2023.1232177}}


\bibitem[Brown(2023)]%
        {brown2023allocating}
\bibfield{author}{\bibinfo{person}{Ian Brown}.} \bibinfo{year}{2023}\natexlab{}.
\newblock \showarticletitle{Allocating accountability in AI supply chains}.
\newblock \bibinfo{journal}{\emph{Ada Lovelace Institute}} (\bibinfo{year}{2023}).
\newblock


\bibitem[Carro(2024)]%
        {Carro_2024}
\bibfield{author}{\bibinfo{person}{María~Victoria Carro}.} \bibinfo{year}{2024}\natexlab{}.
\newblock \showarticletitle{Flattering to Deceive: The Impact of Sycophantic Behavior on User Trust in Large Language Model}.
\newblock \bibinfo{journal}{\emph{arXiv Preprint}} \bibinfo{number}{arXiv:2412.02802} (\bibinfo{date}{Dec.} \bibinfo{year}{2024}).
\newblock
\href{https://doi.org/10.48550/arXiv.2412.02802}{doi:\nolinkurl{10.48550/arXiv.2412.02802}}
\newblock
\shownote{arXiv:2412.02802 [cs]}.


\bibitem[Chaudhary and Penn(2024)]%
        {Chaudhary_Penn_2024}
\bibfield{author}{\bibinfo{person}{Yaqub Chaudhary} {and} \bibinfo{person}{Jonnie Penn}.} \bibinfo{year}{2024}\natexlab{}.
\newblock \showarticletitle{Beware the Intention Economy: Collection and Commodification of Intent via Large Language Models}.
\newblock \bibinfo{journal}{\emph{Harvard Data Science Review}} \bibinfo{number}{Special Issue 5} (\bibinfo{date}{Dec.} \bibinfo{year}{2024}).
\newblock
\showISSN{2644-2353, 2688-8513}
\href{https://doi.org/10.1162/99608f92.21e6bbaa}{doi:\nolinkurl{10.1162/99608f92.21e6bbaa}}


\bibitem[Cheong et~al\mbox{.}(2024)]%
        {cheong2024small}
\bibfield{author}{\bibinfo{person}{Jiaee Cheong}, \bibinfo{person}{Micol Spitale}, {and} \bibinfo{person}{Hatice Gunes}.} \bibinfo{year}{2024}\natexlab{}.
\newblock \showarticletitle{Small but fair! fairness for multimodal human-human and robot-human mental wellbeing coaching}.
\newblock \bibinfo{journal}{\emph{arXiv preprint arXiv:2407.01562}} (\bibinfo{year}{2024}).
\newblock


\bibitem[Cobbe et~al\mbox{.}(2023)]%
        {cobbe2023understanding}
\bibfield{author}{\bibinfo{person}{Jennifer Cobbe}, \bibinfo{person}{Michael Veale}, {and} \bibinfo{person}{Jatinder Singh}.} \bibinfo{year}{2023}\natexlab{}.
\newblock \showarticletitle{Understanding accountability in algorithmic supply chains}. In \bibinfo{booktitle}{\emph{Proceedings of the 2023 ACM Conference on Fairness, Accountability, and Transparency}}. \bibinfo{pages}{1186--1197}.
\newblock


\bibitem[Costanza-Chock(2020)]%
        {costanza2020design}
\bibfield{author}{\bibinfo{person}{Sasha Costanza-Chock}.} \bibinfo{year}{2020}\natexlab{}.
\newblock \bibinfo{booktitle}{\emph{Design justice: Community-led practices to build the worlds we need}}.
\newblock \bibinfo{publisher}{The MIT Press}, \bibinfo{address}{Cambridge, Massachusetts}.
\newblock


\bibitem[Dehnert(2024)]%
        {dehnert2024ability}
\bibfield{author}{\bibinfo{person}{Marco Dehnert}.} \bibinfo{year}{2024}\natexlab{}.
\newblock \showarticletitle{Ability and Disability: Social Robots and Accessibility, Disability Justice, and the Socially Constructed Normal Body}.
\newblock \bibinfo{journal}{\emph{The De Gruyter Handbook of Robots in Society and Culture}}  \bibinfo{volume}{3} (\bibinfo{year}{2024}), \bibinfo{pages}{429}.
\newblock


\bibitem[Doğan et~al\mbox{.}(2023)]%
        {10.3389/frobt.2022.937772}
\bibfield{author}{\bibinfo{person}{Fethiye~Irmak Doğan}, \bibinfo{person}{Gaspar~I. Melsión}, {and} \bibinfo{person}{Iolanda Leite}.} \bibinfo{year}{2023}\natexlab{}.
\newblock \showarticletitle{Leveraging explainability for understanding object descriptions in ambiguous 3D environments}.
\newblock \bibinfo{journal}{\emph{Frontiers in Robotics and AI}}  \bibinfo{volume}{Volume 9 - 2022} (\bibinfo{year}{2023}).
\newblock
\showISSN{2296-9144}
\href{https://doi.org/10.3389/frobt.2022.937772}{doi:\nolinkurl{10.3389/frobt.2022.937772}}


\bibitem[Doğan et~al\mbox{.}(2025)]%
        {11127826}
\bibfield{author}{\bibinfo{person}{Fethiye~Irmak Doğan}, \bibinfo{person}{Umut Ozyurt}, \bibinfo{person}{Gizem Cinar}, {and} \bibinfo{person}{Hatice Gunes}.} \bibinfo{year}{2025}\natexlab{}.
\newblock \showarticletitle{GRACE: Generating Socially Appropriate Robot Actions Leveraging LLMs and Human Explanations}. In \bibinfo{booktitle}{\emph{2025 IEEE International Conference on Robotics and Automation (ICRA)}}. \bibinfo{pages}{4330--4336}.
\newblock
\href{https://doi.org/10.1109/ICRA55743.2025.11127826}{doi:\nolinkurl{10.1109/ICRA55743.2025.11127826}}


\bibitem[Doğan et~al\mbox{.}(2022)]%
        {9889368}
\bibfield{author}{\bibinfo{person}{Fethiye~Irmak Doğan}, \bibinfo{person}{Ilaria Torre}, {and} \bibinfo{person}{Iolanda Leite}.} \bibinfo{year}{2022}\natexlab{}.
\newblock \showarticletitle{Asking Follow-Up Clarifications to Resolve Ambiguities in Human-Robot Conversation}. In \bibinfo{booktitle}{\emph{2022 17th ACM/IEEE International Conference on Human-Robot Interaction (HRI)}}. \bibinfo{pages}{461--469}.
\newblock
\href{https://doi.org/10.1109/HRI53351.2022.9889368}{doi:\nolinkurl{10.1109/HRI53351.2022.9889368}}


\bibitem[Edmonds et~al\mbox{.}(2019)]%
        {edmonds2019tale}
\bibfield{author}{\bibinfo{person}{Mark Edmonds}, \bibinfo{person}{Feng Gao}, \bibinfo{person}{Hangxin Liu}, \bibinfo{person}{Xu Xie}, \bibinfo{person}{Siyuan Qi}, \bibinfo{person}{Brandon Rothrock}, \bibinfo{person}{Yixin Zhu}, \bibinfo{person}{Ying~Nian Wu}, \bibinfo{person}{Hongjing Lu}, {and} \bibinfo{person}{Song-Chun Zhu}.} \bibinfo{year}{2019}\natexlab{}.
\newblock \showarticletitle{A tale of two explanations: Enhancing human trust by explaining robot behavior}.
\newblock \bibinfo{journal}{\emph{Science Robotics}} \bibinfo{volume}{4}, \bibinfo{number}{37} (\bibinfo{year}{2019}).
\newblock


\bibitem[Ehsan and Riedl(2024)]%
        {ehsan2024explainability}
\bibfield{author}{\bibinfo{person}{Upol Ehsan} {and} \bibinfo{person}{Mark~O Riedl}.} \bibinfo{year}{2024}\natexlab{}.
\newblock \showarticletitle{Explainability pitfalls: Beyond dark patterns in explainable AI}.
\newblock \bibinfo{journal}{\emph{Patterns}} \bibinfo{volume}{5}, \bibinfo{number}{6} (\bibinfo{year}{2024}).
\newblock


\bibitem[Felkner et~al\mbox{.}(2024)]%
        {Felkner_Chang_Jang_May_2024}
\bibfield{author}{\bibinfo{person}{Virginia~K. Felkner}, \bibinfo{person}{Ho-Chun~Herbert Chang}, \bibinfo{person}{Eugene Jang}, {and} \bibinfo{person}{Jonathan May}.} \bibinfo{year}{2024}\natexlab{}.
\newblock \showarticletitle{WinoQueer: A Community-in-the-Loop Benchmark for Anti-LGBTQ+ Bias in Large Language Models}.
\newblock  \bibinfo{number}{arXiv:2306.15087} (\bibinfo{date}{Oct.} \bibinfo{year}{2024}).
\newblock
\href{https://doi.org/10.48550/arXiv.2306.15087}{doi:\nolinkurl{10.48550/arXiv.2306.15087}}
\newblock
\shownote{arXiv:2306.15087 [cs]}.


\bibitem[Frennert et~al\mbox{.}(2024)]%
        {Frennert_Persson_Skavron_2024}
\bibfield{author}{\bibinfo{person}{Susanne Frennert}, \bibinfo{person}{Johanna Persson}, {and} \bibinfo{person}{Sarah Skavron}.} \bibinfo{year}{2024}\natexlab{}.
\newblock \showarticletitle{A Critical Narrative Review of Assistive Robotics and Call for a Systems and User-Centered Approaches to Enhance Quality of Life of Individuals with Disabilities}. In \bibinfo{booktitle}{\emph{Adjunct Proceedings of the 2024 Nordic Conference on Human-Computer Interaction}}. \bibinfo{publisher}{ACM}, \bibinfo{address}{Uppsala Sweden}, \bibinfo{pages}{1–11}.
\newblock
\showISBNx{9798400709654}
\href{https://doi.org/10.1145/3677045.3685495}{doi:\nolinkurl{10.1145/3677045.3685495}}


\bibitem[Guerdan et~al\mbox{.}(2021)]%
        {9607734}
\bibfield{author}{\bibinfo{person}{Luke Guerdan}, \bibinfo{person}{Alex Raymond}, {and} \bibinfo{person}{Hatice Gunes}.} \bibinfo{year}{2021}\natexlab{}.
\newblock \showarticletitle{Toward Affective XAI: Facial Affect Analysis for Understanding Explainable Human-AI Interactions}. In \bibinfo{booktitle}{\emph{2021 IEEE/CVF International Conference on Computer Vision Workshops (ICCVW)}}. \bibinfo{pages}{3789--3798}.
\newblock
\href{https://doi.org/10.1109/ICCVW54120.2021.00423}{doi:\nolinkurl{10.1109/ICCVW54120.2021.00423}}


\bibitem[Hanschmann et~al\mbox{.}(2024)]%
        {Hanschmann_Gnewuch_Maedche_2024}
\bibfield{author}{\bibinfo{person}{Leon Hanschmann}, \bibinfo{person}{Ulrich Gnewuch}, {and} \bibinfo{person}{Alexander Maedche}.} \bibinfo{year}{2024}\natexlab{}.
\newblock \showarticletitle{Saleshat: A LLM-Based Social Robot for Human-Like Sales Conversations}. In \bibinfo{booktitle}{\emph{Chatbot Research and Design}}, \bibfield{editor}{\bibinfo{person}{Asbjørn Følstad}, \bibinfo{person}{Theo Araujo}, \bibinfo{person}{Symeon Papadopoulos}, \bibinfo{person}{Effie L.-C. Law}, \bibinfo{person}{Ewa Luger}, \bibinfo{person}{Morten Goodwin}, \bibinfo{person}{Sebastian Hobert}, {and} \bibinfo{person}{Petter~Bae Brandtzaeg}} (Eds.). \bibinfo{publisher}{Springer Nature Switzerland}, \bibinfo{address}{Cham}, \bibinfo{pages}{61–76}.
\newblock
\showISBNx{978-3-031-54975-5}
\href{https://doi.org/10.1007/978-3-031-54975-5_4}{doi:\nolinkurl{10.1007/978-3-031-54975-5_4}}


\bibitem[Hanson(2025)]%
        {hanson2025indigenous}
\bibfield{author}{\bibinfo{person}{Zachary~Arao Hanson}.} \bibinfo{year}{2025}\natexlab{}.
\newblock \showarticletitle{Indigenous (Mis) Representation in Emerging LLM Research Methodologies}.
\newblock \bibinfo{journal}{\emph{UC Riverside Undergraduate Research Journal}} \bibinfo{volume}{19}, \bibinfo{number}{1} (\bibinfo{year}{2025}).
\newblock


\bibitem[Hila(2025)]%
        {hila2025epistemological}
\bibfield{author}{\bibinfo{person}{Angjelin Hila}.} \bibinfo{year}{2025}\natexlab{}.
\newblock \showarticletitle{The epistemological consequences of large language models: rethinking collective intelligence and institutional knowledge}.
\newblock \bibinfo{journal}{\emph{AI \& SOCIETY}} (\bibinfo{year}{2025}), \bibinfo{pages}{1--19}.
\newblock


\bibitem[Hu et~al\mbox{.}(2022)]%
        {hu2022lora}
\bibfield{author}{\bibinfo{person}{Edward~J Hu}, \bibinfo{person}{Yelong Shen}, \bibinfo{person}{Phillip Wallis}, \bibinfo{person}{Zeyuan Allen-Zhu}, \bibinfo{person}{Yuanzhi Li}, \bibinfo{person}{Shean Wang}, \bibinfo{person}{Lu Wang}, \bibinfo{person}{Weizhu Chen}, {et~al\mbox{.}}} \bibinfo{year}{2022}\natexlab{}.
\newblock \showarticletitle{Lora: Low-rank adaptation of large language models.}
\newblock \bibinfo{journal}{\emph{ICLR}} \bibinfo{volume}{1}, \bibinfo{number}{2} (\bibinfo{year}{2022}), \bibinfo{pages}{3}.
\newblock


\bibitem[Irfan et~al\mbox{.}(2025)]%
        {Irfan_Kuoppamäki_Hosseini_Skantze_2025}
\bibfield{author}{\bibinfo{person}{Bahar Irfan}, \bibinfo{person}{Sanna Kuoppamäki}, \bibinfo{person}{Aida Hosseini}, {and} \bibinfo{person}{Gabriel Skantze}.} \bibinfo{year}{2025}\natexlab{}.
\newblock \showarticletitle{Between reality and delusion: challenges of applying large language models to companion robots for open-domain dialogues with older adults}.
\newblock \bibinfo{journal}{\emph{Autonomous Robots}} \bibinfo{volume}{49}, \bibinfo{number}{1} (\bibinfo{date}{March} \bibinfo{year}{2025}), \bibinfo{pages}{9}.
\newblock
\showISSN{1573-7527}
\href{https://doi.org/10.1007/s10514-025-10190-y}{doi:\nolinkurl{10.1007/s10514-025-10190-y}}


\bibitem[Irfan et~al\mbox{.}(2024)]%
        {Irfan_Kuoppamäki_Skantze_2024}
\bibfield{author}{\bibinfo{person}{Bahar Irfan}, \bibinfo{person}{Sanna Kuoppamäki}, {and} \bibinfo{person}{Gabriel Skantze}.} \bibinfo{year}{2024}\natexlab{}.
\newblock \showarticletitle{Recommendations for designing conversational companion robots with older adults through foundation models}.
\newblock \bibinfo{journal}{\emph{Frontiers in Robotics and AI}}  \bibinfo{volume}{11} (\bibinfo{date}{May} \bibinfo{year}{2024}).
\newblock
\showISSN{2296-9144}
\href{https://doi.org/10.3389/frobt.2024.1363713}{doi:\nolinkurl{10.3389/frobt.2024.1363713}}


\bibitem[Jokinen and Wilcock(2024)]%
        {Jokinen_Wilcock_2024}
\bibfield{author}{\bibinfo{person}{Kristiina Jokinen} {and} \bibinfo{person}{Graham Wilcock}.} \bibinfo{year}{2024}\natexlab{}.
\newblock \showarticletitle{Exploring a Japanese Cooking Database}. In \bibinfo{booktitle}{\emph{Companion of the 2024 ACM/IEEE International Conference on Human-Robot Interaction}}. \bibinfo{publisher}{ACM}, \bibinfo{address}{Boulder CO USA}, \bibinfo{pages}{578–582}.
\newblock
\showISBNx{979-8-4007-0323-2}
\href{https://doi.org/10.1145/3610978.3640622}{doi:\nolinkurl{10.1145/3610978.3640622}}


\bibitem[Khanna and Li(2025)]%
        {khanna2025invisible}
\bibfield{author}{\bibinfo{person}{Saurabh Khanna} {and} \bibinfo{person}{Xinxu Li}.} \bibinfo{year}{2025}\natexlab{}.
\newblock \showarticletitle{Invisible Languages of the LLM Universe}.
\newblock \bibinfo{journal}{\emph{arXiv preprint arXiv:2510.11557}} (\bibinfo{year}{2025}).
\newblock


\bibitem[Kotek et~al\mbox{.}(2023)]%
        {Kotek_Dockum_Sun_2023}
\bibfield{author}{\bibinfo{person}{Hadas Kotek}, \bibinfo{person}{Rikker Dockum}, {and} \bibinfo{person}{David Sun}.} \bibinfo{year}{2023}\natexlab{}.
\newblock \showarticletitle{Gender bias and stereotypes in Large Language Models}. In \bibinfo{booktitle}{\emph{Proceedings of The ACM Collective Intelligence Conference}} \emph{(\bibinfo{series}{CI ’23})}. \bibinfo{publisher}{Association for Computing Machinery}, \bibinfo{address}{New York, NY, USA}, \bibinfo{pages}{12–24}.
\newblock
\showISBNx{979-8-4007-0113-9}
\href{https://doi.org/10.1145/3582269.3615599}{doi:\nolinkurl{10.1145/3582269.3615599}}


\bibitem[Kunz and {\'{O}}~h{\'{E}}igeartaigh(2020)]%
        {kunz2020}
\bibfield{author}{\bibinfo{person}{Michael Kunz} {and} \bibinfo{person}{Seán {\'{O}}~h{\'{E}}igeartaigh}.} \bibinfo{year}{2020}\natexlab{}.
\newblock \showarticletitle{Artificial Intelligence and Robotization}.
\newblock In \bibinfo{booktitle}{\emph{Oxford Handbook on the International Law of Global Security}}, \bibfield{editor}{\bibinfo{person}{Robin Geiß} {and} \bibinfo{person}{Nils Melzer}} (Eds.). \bibinfo{publisher}{Oxford University Press}, \bibinfo{address}{Oxford}.
\newblock
\newblock
\shownote{Forthcoming}.


\bibitem[Lammert(2026)]%
        {10.1007/978-3-032-08333-3_11}
\bibfield{author}{\bibinfo{person}{Olesja Lammert}.} \bibinfo{year}{2026}\natexlab{}.
\newblock \showarticletitle{Can AI Regulate Your Emotions? An Empirical Investigation of the Influence of AI Explanations and Emotion Regulation on Human Decision-Making Factors}. In \bibinfo{booktitle}{\emph{Explainable Artificial Intelligence}}, \bibfield{editor}{\bibinfo{person}{Riccardo Guidotti}, \bibinfo{person}{Ute Schmid}, {and} \bibinfo{person}{Luca Longo}} (Eds.). \bibinfo{publisher}{Springer Nature Switzerland}, \bibinfo{address}{Cham}, \bibinfo{pages}{225--248}.
\newblock
\showISBNx{978-3-032-08333-3}


\bibitem[Leusmann et~al\mbox{.}(2024)]%
        {leusmann2024comparing}
\bibfield{author}{\bibinfo{person}{Jan Leusmann}, \bibinfo{person}{Chao Wang}, {and} \bibinfo{person}{Sven Mayer}.} \bibinfo{year}{2024}\natexlab{}.
\newblock \showarticletitle{Comparing Rule-based and LLM-based Methods to Enable Active Robot Assistant Conversations}.
\newblock \bibinfo{journal}{\emph{Proceedings of the CUI@ CHI 2024: Building Trust in CUIs--From Design to Deployment}} (\bibinfo{year}{2024}), \bibinfo{pages}{05--11}.
\newblock


\bibitem[Markelius(2025)]%
        {Markelius}
\bibfield{author}{\bibinfo{person}{Alva Markelius}.} \bibinfo{year}{2025}\natexlab{}.
\newblock \bibinfo{booktitle}{\emph{An Empirical Design Justice Approach to Identifying Ethical Considerations in the Intersection of Large Language Models and Social Robotics}}.
\newblock \bibinfo{publisher}{Oxford University Press}.
\newblock
\showISBNx{978-0-19-894521-5}
\href{https://doi.org/10.1093/9780198945215.003.0013}{doi:\nolinkurl{10.1093/9780198945215.003.0013}}


\bibitem[Markelius et~al\mbox{.}(2025)]%
        {markelius2025stakeholder}
\bibfield{author}{\bibinfo{person}{Alva Markelius}, \bibinfo{person}{Julie Bailey}, \bibinfo{person}{Jenny~L Gibson}, {and} \bibinfo{person}{Hatice Gunes}.} \bibinfo{year}{2025}\natexlab{}.
\newblock \showarticletitle{Stakeholder Perspectives on Whether and How Social Robots Can Support Mediation and Advocacy for Higher Education Students with Disabilities}.
\newblock \bibinfo{journal}{\emph{arXiv preprint arXiv:2503.16499}} (\bibinfo{year}{2025}).
\newblock


\bibitem[Markelius and Gunes(2025)]%
        {markscoping}
\bibfield{author}{\bibinfo{person}{Alva Markelius} {and} \bibinfo{person}{Hatice Gunes}.} \bibinfo{year}{2025}\natexlab{}.
\newblock \showarticletitle{Social Robotics and Large Language Models for Disability: A Scoping Review}.
\newblock \bibinfo{journal}{\emph{Research Square Preprint}} (\bibinfo{year}{2025}).
\newblock
\href{https://doi.org/10.21203/rs.3.rs-7260703/v1}{doi:\nolinkurl{10.21203/rs.3.rs-7260703/v1}}


\bibitem[Martins et~al\mbox{.}(2019)]%
        {martins2019user}
\bibfield{author}{\bibinfo{person}{Gon{\c{c}}alo~S Martins}, \bibinfo{person}{Lu{\'\i}s Santos}, {and} \bibinfo{person}{Jorge Dias}.} \bibinfo{year}{2019}\natexlab{}.
\newblock \showarticletitle{User-adaptive interaction in social robots: A survey focusing on non-physical interaction}.
\newblock \bibinfo{journal}{\emph{International Journal of Social Robotics}} \bibinfo{volume}{11}, \bibinfo{number}{1} (\bibinfo{year}{2019}), \bibinfo{pages}{185--205}.
\newblock


\bibitem[Masters et~al\mbox{.}(2025)]%
        {10.5555/3709347.3743971}
\bibfield{author}{\bibinfo{person}{Peta Masters}, \bibinfo{person}{Daniel Gallagher}, \bibinfo{person}{Luc Moreau}, {and} \bibinfo{person}{Mor Vered}.} \bibinfo{year}{2025}\natexlab{}.
\newblock \showarticletitle{Rethinking Explainable AI: Explanations can be Deceiving}. In \bibinfo{booktitle}{\emph{Proceedings of the 24th International Conference on Autonomous Agents and Multiagent Systems}} (Detroit, MI, USA) \emph{(\bibinfo{series}{AAMAS '25})}. \bibinfo{publisher}{International Foundation for Autonomous Agents and Multiagent Systems}, \bibinfo{address}{Richland, SC}, \bibinfo{pages}{2663–2665}.
\newblock
\showISBNx{9798400714269}


\bibitem[Mishra et~al\mbox{.}(2023)]%
        {Mishra_Verdonschot_Hagoort_Skantze_2023}
\bibfield{author}{\bibinfo{person}{Chinmaya Mishra}, \bibinfo{person}{Rinus Verdonschot}, \bibinfo{person}{Peter Hagoort}, {and} \bibinfo{person}{Gabriel Skantze}.} \bibinfo{year}{2023}\natexlab{}.
\newblock \showarticletitle{Real-time emotion generation in human-robot dialogue using large language models}.
\newblock \bibinfo{journal}{\emph{Frontiers in Robotics and AI}}  \bibinfo{volume}{10} (\bibinfo{date}{Dec.} \bibinfo{year}{2023}).
\newblock
\showISSN{2296-9144}
\href{https://doi.org/10.3389/frobt.2023.1271610}{doi:\nolinkurl{10.3389/frobt.2023.1271610}}


\bibitem[Murali et~al\mbox{.}(2023)]%
        {10.1145/3544549.3585602}
\bibfield{author}{\bibinfo{person}{Prasanth Murali}, \bibinfo{person}{Ian Steenstra}, \bibinfo{person}{Hye~Sun Yun}, \bibinfo{person}{Ameneh Shamekhi}, {and} \bibinfo{person}{Timothy Bickmore}.} \bibinfo{year}{2023}\natexlab{}.
\newblock \showarticletitle{Improving Multiparty Interactions with a Robot Using Large Language Models}. In \bibinfo{booktitle}{\emph{Extended Abstracts of the 2023 CHI Conference on Human Factors in Computing Systems}} (Hamburg, Germany) \emph{(\bibinfo{series}{CHI EA '23})}. \bibinfo{publisher}{Association for Computing Machinery}, \bibinfo{address}{New York, NY, USA}, Article \bibinfo{articleno}{175}, \bibinfo{numpages}{8}~pages.
\newblock
\showISBNx{9781450394222}
\href{https://doi.org/10.1145/3544549.3585602}{doi:\nolinkurl{10.1145/3544549.3585602}}


\bibitem[Onorati et~al\mbox{.}(2023)]%
        {onorati2023creating}
\bibfield{author}{\bibinfo{person}{Teresa Onorati}, \bibinfo{person}{{\'A}lvaro Castro-Gonz{\'a}lez}, \bibinfo{person}{Javier~Cruz del Valle}, \bibinfo{person}{Paloma D{\'\i}az}, {and} \bibinfo{person}{Jos{\'e}~Carlos Castillo}.} \bibinfo{year}{2023}\natexlab{}.
\newblock \showarticletitle{Creating personalized verbal human-robot interactions using llm with the robot mini}. In \bibinfo{booktitle}{\emph{International conference on ubiquitous computing and ambient intelligence}}. Springer, \bibinfo{pages}{148--159}.
\newblock


\bibitem[Ostrowski et~al\mbox{.}(2022)]%
        {ostrowski2022ethics}
\bibfield{author}{\bibinfo{person}{Anastasia~K Ostrowski}, \bibinfo{person}{Raechel Walker}, \bibinfo{person}{Madhurima Das}, \bibinfo{person}{Maria Yang}, \bibinfo{person}{Cynthia Breazea}, \bibinfo{person}{Hae~Won Park}, {and} \bibinfo{person}{Aditi Verma}.} \bibinfo{year}{2022}\natexlab{}.
\newblock \showarticletitle{Ethics, equity, \& justice in human-robot interaction: A review and future directions}. In \bibinfo{booktitle}{\emph{2022 31st IEEE International Conference on Robot and Human Interactive Communication (RO-MAN)}}. IEEE, \bibinfo{pages}{969--976}.
\newblock
\href{https://doi.org/10.1109/RO-MAN53752.2022.9900805}{doi:\nolinkurl{10.1109/RO-MAN53752.2022.9900805}}


\bibitem[Padmanabha et~al\mbox{.}(2024)]%
        {Padmanabha_Yuan_Gupta_Karachiwalla_Majidi_Admoni_Erickson_2024}
\bibfield{author}{\bibinfo{person}{Akhil Padmanabha}, \bibinfo{person}{Jessie Yuan}, \bibinfo{person}{Janavi Gupta}, \bibinfo{person}{Zulekha Karachiwalla}, \bibinfo{person}{Carmel Majidi}, \bibinfo{person}{Henny Admoni}, {and} \bibinfo{person}{Zackory Erickson}.} \bibinfo{year}{2024}\natexlab{}.
\newblock \showarticletitle{VoicePilot: Harnessing LLMs as Speech Interfaces for Physically Assistive Robots}. In \bibinfo{booktitle}{\emph{Proceedings of the 37th Annual ACM Symposium on User Interface Software and Technology}} \emph{(\bibinfo{series}{UIST ’24})}. \bibinfo{publisher}{Association for Computing Machinery}, \bibinfo{address}{New York, NY, USA}, \bibinfo{pages}{1–18}.
\newblock
\showISBNx{9798400706288}
\href{https://doi.org/10.1145/3654777.3676401}{doi:\nolinkurl{10.1145/3654777.3676401}}


\bibitem[Palacios~Barea et~al\mbox{.}(2025)]%
        {palacios2025intersection}
\bibfield{author}{\bibinfo{person}{MA Palacios~Barea}, \bibinfo{person}{D Boeren}, {and} \bibinfo{person}{JF Ferreira~Goncalves}.} \bibinfo{year}{2025}\natexlab{}.
\newblock \showarticletitle{At the intersection of humanity and technology: a technofeminist intersectional critical discourse analysis of gender and race biases in the natural language processing model GPT-3}.
\newblock \bibinfo{journal}{\emph{AI \& SOCIETY}} \bibinfo{volume}{40}, \bibinfo{number}{2} (\bibinfo{year}{2025}), \bibinfo{pages}{461--479}.
\newblock


\bibitem[Panesar et~al\mbox{.}(2022)]%
        {9900586}
\bibfield{author}{\bibinfo{person}{Amrita Panesar}, \bibinfo{person}{Fethiye~Irmak Doğan}, {and} \bibinfo{person}{Iolanda Leite}.} \bibinfo{year}{2022}\natexlab{}.
\newblock \showarticletitle{Improving Visual Question Answering by Leveraging Depth and Adapting Explainability}. In \bibinfo{booktitle}{\emph{2022 31st IEEE International Conference on Robot and Human Interactive Communication (RO-MAN)}}. \bibinfo{pages}{252--259}.
\newblock
\href{https://doi.org/10.1109/RO-MAN53752.2022.9900586}{doi:\nolinkurl{10.1109/RO-MAN53752.2022.9900586}}


\bibitem[Paterson(2024)]%
        {paterson2024robot}
\bibfield{author}{\bibinfo{person}{Mark Paterson}.} \bibinfo{year}{2024}\natexlab{}.
\newblock \showarticletitle{Why robot embodiment matters: questions of disability, race and intersectionality in the design of social robots}.
\newblock \bibinfo{journal}{\emph{Medical Humanities}} \bibinfo{volume}{50}, \bibinfo{number}{4} (\bibinfo{year}{2024}), \bibinfo{pages}{694--704}.
\newblock


\bibitem[Policies(2023)]%
        {Whatpale}
\bibfield{author}{\bibinfo{person}{International~Health Policies}.} \bibinfo{year}{2023}\natexlab{}.
\newblock \bibinfo{title}{What Happened-to all Human Beings are Born Free Reflections on a Chatgpt-Experiment}.
\newblock
\urldef\tempurl%
\url{https://www.internationalhealthpolicies.org/featured-article/what-happened-to-all-human-beings-are-born-free-reflections-on-a-chatgpt-experiment/}
\showURL{%
\tempurl}


\bibitem[Rakova et~al\mbox{.}(2023)]%
        {rakova2023terms}
\bibfield{author}{\bibinfo{person}{Bogdana Rakova}, \bibinfo{person}{Renee Shelby}, {and} \bibinfo{person}{Megan Ma}.} \bibinfo{year}{2023}\natexlab{}.
\newblock \showarticletitle{Terms-we-serve-with: Five dimensions for anticipating and repairing algorithmic harm}.
\newblock \bibinfo{journal}{\emph{Big Data \& Society}} \bibinfo{volume}{10}, \bibinfo{number}{2} (\bibinfo{year}{2023}), \bibinfo{pages}{20539517231211553}.
\newblock


\bibitem[Raymond et~al\mbox{.}(2020)]%
        {10.5555/3398761.3398890}
\bibfield{author}{\bibinfo{person}{Alex Raymond}, \bibinfo{person}{Hatice Gunes}, {and} \bibinfo{person}{Amanda Prorok}.} \bibinfo{year}{2020}\natexlab{}.
\newblock \showarticletitle{Culture-Based Explainable Human-Agent Deconfliction}. In \bibinfo{booktitle}{\emph{Proceedings of the 19th International Conference on Autonomous Agents and MultiAgent Systems}} (Auckland, New Zealand) \emph{(\bibinfo{series}{AAMAS '20})}. \bibinfo{publisher}{International Foundation for Autonomous Agents and Multiagent Systems}, \bibinfo{address}{Richland, SC}, \bibinfo{pages}{1107–1115}.
\newblock
\showISBNx{9781450375184}


\bibitem[Rizvi et~al\mbox{.}(2024)]%
        {10.1145/3613904.3642798}
\bibfield{author}{\bibinfo{person}{Naba Rizvi}, \bibinfo{person}{William Wu}, \bibinfo{person}{Mya Bolds}, \bibinfo{person}{Raunak Mondal}, \bibinfo{person}{Andrew Begel}, {and} \bibinfo{person}{Imani N.~S. Munyaka}.} \bibinfo{year}{2024}\natexlab{}.
\newblock \showarticletitle{Are Robots Ready to Deliver Autism Inclusion?: A Critical Review} \emph{(\bibinfo{series}{CHI '24})}. \bibinfo{publisher}{Association for Computing Machinery}, \bibinfo{address}{New York, NY, USA}, Article \bibinfo{articleno}{69}, \bibinfo{numpages}{18}~pages.
\newblock
\showISBNx{9798400703300}
\href{https://doi.org/10.1145/3613904.3642798}{doi:\nolinkurl{10.1145/3613904.3642798}}


\bibitem[Sakai and Nagai(2022)]%
        {sakai2022explainable}
\bibfield{author}{\bibinfo{person}{Tatsuya Sakai} {and} \bibinfo{person}{Takayuki Nagai}.} \bibinfo{year}{2022}\natexlab{}.
\newblock \showarticletitle{Explainable autonomous robots: a survey and perspective}.
\newblock \bibinfo{journal}{\emph{Advanced Robotics}} \bibinfo{volume}{36}, \bibinfo{number}{5-6} (\bibinfo{year}{2022}), \bibinfo{pages}{219--238}.
\newblock


\bibitem[Seaborn(2022)]%
        {seaborn2022not}
\bibfield{author}{\bibinfo{person}{Katie Seaborn}.} \bibinfo{year}{2022}\natexlab{}.
\newblock \showarticletitle{Not Only WEIRD but UNCANNY: A Critical Intersectional Review of Social Robotics Research}.
\newblock \bibinfo{journal}{\emph{OSF}} (\bibinfo{year}{2022}).
\newblock


\bibitem[Seaborn et~al\mbox{.}(2023)]%
        {seaborn2023not}
\bibfield{author}{\bibinfo{person}{Katie Seaborn}, \bibinfo{person}{Giulia Barbareschi}, {and} \bibinfo{person}{Shruti Chandra}.} \bibinfo{year}{2023}\natexlab{}.
\newblock \showarticletitle{Not only WEIRD but “uncanny”? A systematic review of diversity in human--robot interaction research}.
\newblock \bibinfo{journal}{\emph{International Journal of Social Robotics}} \bibinfo{volume}{15}, \bibinfo{number}{11} (\bibinfo{year}{2023}), \bibinfo{pages}{1841--1870}.
\newblock


\bibitem[Setchi et~al\mbox{.}(2020)]%
        {setchi2020explainable}
\bibfield{author}{\bibinfo{person}{Rossitza Setchi}, \bibinfo{person}{Maryam~Banitalebi Dehkordi}, {and} \bibinfo{person}{Juwairiya~Siraj Khan}.} \bibinfo{year}{2020}\natexlab{}.
\newblock \showarticletitle{Explainable Robotics in Human-Robot Interactions}.
\newblock \bibinfo{journal}{\emph{Procedia Computer Science}}  \bibinfo{volume}{176} (\bibinfo{year}{2020}), \bibinfo{pages}{3057--3066}.
\newblock


\bibitem[Sevilla-Salcedo et~al\mbox{.}(2023)]%
        {sevilla2023using}
\bibfield{author}{\bibinfo{person}{Javier Sevilla-Salcedo}, \bibinfo{person}{Enrique Fern{\'a}dez-Rodicio}, \bibinfo{person}{Laura Mart{\'\i}n-Galv{\'a}n}, \bibinfo{person}{{\'A}lvaro Castro-Gonz{\'a}lez}, \bibinfo{person}{Jos{\'e}~C Castillo}, {and} \bibinfo{person}{Miguel~A Salichs}.} \bibinfo{year}{2023}\natexlab{}.
\newblock \showarticletitle{Using large language models to shape social robots’ speech}.
\newblock  (\bibinfo{year}{2023}).
\newblock


\bibitem[Sherer et~al\mbox{.}(2025)]%
        {Sherer_McPherson_Mohanty_Santé_Gandolfi_Romeo_Suglia_2025}
\bibfield{author}{\bibinfo{person}{Jeffrey Sherer}, \bibinfo{person}{Robbie McPherson}, \bibinfo{person}{Sattwik Mohanty}, \bibinfo{person}{Guilhem Santé}, \bibinfo{person}{Greta Gandolfi}, \bibinfo{person}{Marta Romeo}, {and} \bibinfo{person}{Alessandro Suglia}.} \bibinfo{year}{2025}\natexlab{}.
\newblock \showarticletitle{Follow Me: A Study on the Dynamics of Alignment Between Humans and LLM-Based Social Robots}. In \bibinfo{booktitle}{\emph{Social Robotics}}, \bibfield{editor}{\bibinfo{person}{Oskar Palinko}, \bibinfo{person}{Leon Bodenhagen}, \bibinfo{person}{John-John Cabibihan}, \bibinfo{person}{Kerstin Fischer}, \bibinfo{person}{Selma Šabanović}, \bibinfo{person}{Katie Winkle}, \bibinfo{person}{Laxmidhar Behera}, \bibinfo{person}{Shuzhi~Sam Ge}, \bibinfo{person}{Dimitrios Chrysostomou}, \bibinfo{person}{Wanyue Jiang}, {and} \bibinfo{person}{Hongsheng He}} (Eds.). \bibinfo{publisher}{Springer Nature}, \bibinfo{address}{Singapore}, \bibinfo{pages}{487–496}.
\newblock
\showISBNx{978-981-96-3519-1}
\href{https://doi.org/10.1007/978-981-96-3519-1_44}{doi:\nolinkurl{10.1007/978-981-96-3519-1_44}}


\bibitem[Shi et~al\mbox{.}(2024)]%
        {Shi_Landrum_O’Connell_Kian_Pinto-Alva_Shrestha_Zhu_Matarić_2024}
\bibfield{author}{\bibinfo{person}{Zhonghao Shi}, \bibinfo{person}{Ellen Landrum}, \bibinfo{person}{Amy O’Connell}, \bibinfo{person}{Mina Kian}, \bibinfo{person}{Leticia Pinto-Alva}, \bibinfo{person}{Kaleen Shrestha}, \bibinfo{person}{Xiaoyuan Zhu}, {and} \bibinfo{person}{Maja~J. Matarić}.} \bibinfo{year}{2024}\natexlab{}.
\newblock \showarticletitle{How Can Large Language Models Enable Better Socially Assistive Human-Robot Interaction: A Brief Survey}.
\newblock \bibinfo{journal}{\emph{Proceedings of the AAAI Symposium Series}} \bibinfo{volume}{3}, \bibinfo{number}{11} (\bibinfo{date}{May} \bibinfo{year}{2024}), \bibinfo{pages}{401–404}.
\newblock
\showISSN{2994-4317}
\href{https://doi.org/10.1609/aaaiss.v3i1.31245}{doi:\nolinkurl{10.1609/aaaiss.v3i1.31245}}


\bibitem[Shojaee et~al\mbox{.}(2025)]%
        {shojaee2025illusion}
\bibfield{author}{\bibinfo{person}{Parshin Shojaee}, \bibinfo{person}{Iman Mirzadeh}, \bibinfo{person}{Keivan Alizadeh}, \bibinfo{person}{Maxwell Horton}, \bibinfo{person}{Samy Bengio}, {and} \bibinfo{person}{Mehrdad Farajtabar}.} \bibinfo{year}{2025}\natexlab{}.
\newblock \showarticletitle{The illusion of thinking: Understanding the strengths and limitations of reasoning models via the lens of problem complexity}.
\newblock \bibinfo{journal}{\emph{arXiv preprint arXiv:2506.06941}} (\bibinfo{year}{2025}).
\newblock


\bibitem[Siau and Wang(2018)]%
        {siau2018building}
\bibfield{author}{\bibinfo{person}{Keng Siau} {and} \bibinfo{person}{Weiyu Wang}.} \bibinfo{year}{2018}\natexlab{}.
\newblock \showarticletitle{Building trust in artificial intelligence, machine learning, and robotics}.
\newblock \bibinfo{journal}{\emph{Cutter Business Technology Journal}} \bibinfo{volume}{31}, \bibinfo{number}{2} (\bibinfo{year}{2018}), \bibinfo{pages}{47--53}.
\newblock


\bibitem[Spitale et~al\mbox{.}(2025)]%
        {10.1145/3712265}
\bibfield{author}{\bibinfo{person}{Micol Spitale}, \bibinfo{person}{Minja Axelsson}, {and} \bibinfo{person}{Hatice Gunes}.} \bibinfo{year}{2025}\natexlab{}.
\newblock \showarticletitle{VITA: A Multi-Modal LLM-Based System for Longitudinal, Autonomous and Adaptive Robotic Mental Well-Being Coaching}.
\newblock \bibinfo{journal}{\emph{J. Hum.-Robot Interact.}} \bibinfo{volume}{14}, \bibinfo{number}{2}, Article \bibinfo{articleno}{38} (\bibinfo{date}{March} \bibinfo{year}{2025}), \bibinfo{numpages}{28}~pages.
\newblock
\href{https://doi.org/10.1145/3712265}{doi:\nolinkurl{10.1145/3712265}}


\bibitem[Sridharan and Meadows(2019)]%
        {sridharan2019towards}
\bibfield{author}{\bibinfo{person}{Mohan Sridharan} {and} \bibinfo{person}{Ben Meadows}.} \bibinfo{year}{2019}\natexlab{}.
\newblock \showarticletitle{Towards a Theory of Explanations for Human--Robot Collaboration}.
\newblock \bibinfo{journal}{\emph{KI-K{\"u}nstliche Intelligenz}} \bibinfo{volume}{33}, \bibinfo{number}{4} (\bibinfo{year}{2019}), \bibinfo{pages}{331--342}.
\newblock


\bibitem[Stange et~al\mbox{.}(2022)]%
        {10.3389/frai.2022.866920}
\bibfield{author}{\bibinfo{person}{Sonja Stange}, \bibinfo{person}{Teena Hassan}, \bibinfo{person}{Florian Schröder}, \bibinfo{person}{Jacqueline Konkol}, {and} \bibinfo{person}{Stefan Kopp}.} \bibinfo{year}{2022}\natexlab{}.
\newblock \showarticletitle{Self-Explaining Social Robots: An Explainable Behavior Generation Architecture for Human-Robot Interaction}.
\newblock \bibinfo{journal}{\emph{Frontiers in Artificial Intelligence}}  \bibinfo{volume}{Volume 5 - 2022} (\bibinfo{year}{2022}).
\newblock
\showISSN{2624-8212}
\href{https://doi.org/10.3389/frai.2022.866920}{doi:\nolinkurl{10.3389/frai.2022.866920}}


\bibitem[Voultsiou et~al\mbox{.}(2025)]%
        {Voultsiou_Vrochidou_Moussiades_Papakostas_2025}
\bibfield{author}{\bibinfo{person}{Evdokia Voultsiou}, \bibinfo{person}{Eleni Vrochidou}, \bibinfo{person}{Lefteris Moussiades}, {and} \bibinfo{person}{George~A. Papakostas}.} \bibinfo{year}{2025}\natexlab{}.
\newblock \showarticletitle{The potential of Large Language Models for social robots in special education}.
\newblock \bibinfo{journal}{\emph{Progress in Artificial Intelligence}} (\bibinfo{date}{Feb.} \bibinfo{year}{2025}).
\newblock
\showISSN{2192-6360}
\href{https://doi.org/10.1007/s13748-025-00363-2}{doi:\nolinkurl{10.1007/s13748-025-00363-2}}


\bibitem[Wang et~al\mbox{.}(2024a)]%
        {Wang_Hasler_Tanneberg_Ocker_Joublin_Ceravola_Deigmoeller_Gienger_2024}
\bibfield{author}{\bibinfo{person}{Chao Wang}, \bibinfo{person}{Stephan Hasler}, \bibinfo{person}{Daniel Tanneberg}, \bibinfo{person}{Felix Ocker}, \bibinfo{person}{Frank Joublin}, \bibinfo{person}{Antonello Ceravola}, \bibinfo{person}{Joerg Deigmoeller}, {and} \bibinfo{person}{Michael Gienger}.} \bibinfo{year}{2024}\natexlab{a}.
\newblock \showarticletitle{LaMI: Large Language Models for Multi-Modal Human-Robot Interaction}. In \bibinfo{booktitle}{\emph{Extended Abstracts of the CHI Conference on Human Factors in Computing Systems}} \emph{(\bibinfo{series}{CHI EA ’24})}. \bibinfo{publisher}{Association for Computing Machinery}, \bibinfo{address}{New York, NY, USA}, \bibinfo{pages}{1–10}.
\newblock
\showISBNx{979-8-4007-0331-7}
\href{https://doi.org/10.1145/3613905.3651029}{doi:\nolinkurl{10.1145/3613905.3651029}}


\bibitem[Wang et~al\mbox{.}(2024b)]%
        {Wang_Reisert_Nichols_Gomez_2024}
\bibfield{author}{\bibinfo{person}{Zining Wang}, \bibinfo{person}{Paul Reisert}, \bibinfo{person}{Eric Nichols}, {and} \bibinfo{person}{Randy Gomez}.} \bibinfo{year}{2024}\natexlab{b}.
\newblock \showarticletitle{Ain’t Misbehavin’ - Using LLMs to Generate Expressive Robot Behavior in Conversations with the Tabletop Robot Haru}. In \bibinfo{booktitle}{\emph{Companion of the 2024 ACM/IEEE International Conference on Human-Robot Interaction}} \emph{(\bibinfo{series}{HRI ’24})}. \bibinfo{publisher}{Association for Computing Machinery}, \bibinfo{address}{New York, NY, USA}, \bibinfo{pages}{1105–1109}.
\newblock
\showISBNx{979-8-4007-0323-2}
\href{https://doi.org/10.1145/3610978.3640562}{doi:\nolinkurl{10.1145/3610978.3640562}}


\bibitem[Williams(2021)]%
        {Williams_2021}
\bibfield{author}{\bibinfo{person}{Rua~M. Williams}.} \bibinfo{year}{2021}\natexlab{}.
\newblock \showarticletitle{I, Misfit: Empty Fortresses, Social Robots, and Peculiar Relations in Autism Research}.
\newblock \bibinfo{journal}{\emph{Techné: Research in Philosophy and Technology}} \bibinfo{volume}{25}, \bibinfo{number}{3} (\bibinfo{date}{Nov.} \bibinfo{year}{2021}), \bibinfo{pages}{451–478}.
\newblock
\href{https://doi.org/10.5840/techne20211019147}{doi:\nolinkurl{10.5840/techne20211019147}}


\bibitem[Williams et~al\mbox{.}(2024)]%
        {Williams_Matuszek_Mead_Depalma_2024}
\bibfield{author}{\bibinfo{person}{Tom Williams}, \bibinfo{person}{Cynthia Matuszek}, \bibinfo{person}{Ross Mead}, {and} \bibinfo{person}{Nick Depalma}.} \bibinfo{year}{2024}\natexlab{}.
\newblock \showarticletitle{Scarecrows in Oz: The Use of Large Language Models in HRI}.
\newblock \bibinfo{journal}{\emph{J. Hum.-Robot Interact.}} \bibinfo{volume}{13}, \bibinfo{number}{1} (\bibinfo{date}{Jan.} \bibinfo{year}{2024}), \bibinfo{pages}{1:1--1:11}.
\newblock
\href{https://doi.org/10.1145/3606261}{doi:\nolinkurl{10.1145/3606261}}


\bibitem[Winkle et~al\mbox{.}(2023)]%
        {winkle2023feminist}
\bibfield{author}{\bibinfo{person}{Katie Winkle}, \bibinfo{person}{Donald McMillan}, \bibinfo{person}{Maria Arnelid}, \bibinfo{person}{Katherine Harrison}, \bibinfo{person}{Madeline Balaam}, \bibinfo{person}{Ericka Johnson}, {and} \bibinfo{person}{Iolanda Leite}.} \bibinfo{year}{2023}\natexlab{}.
\newblock \showarticletitle{Feminist human-robot interaction: Disentangling power, principles and practice for better, more ethical HRI}. In \bibinfo{booktitle}{\emph{Proceedings of the 2023 ACM/IEEE international conference on human-robot interaction}}. \bibinfo{pages}{72--82}.
\newblock


\bibitem[Yadollahi et~al\mbox{.}(2025)]%
        {yadollahi2025expectations}
\bibfield{author}{\bibinfo{person}{Elmira Yadollahi}, \bibinfo{person}{Fethiye~Irmak Dogan}, \bibinfo{person}{Yujing Zhang}, \bibinfo{person}{Beatriz Nogueira}, \bibinfo{person}{Tiago Guerreiro}, \bibinfo{person}{Shelly~Levy Tzedek}, {and} \bibinfo{person}{Iolanda Leite}.} \bibinfo{year}{2025}\natexlab{}.
\newblock \showarticletitle{Expectations, Explanations, and Embodiment: Attempts at Robot Failure Recovery}.
\newblock \bibinfo{journal}{\emph{arXiv preprint arXiv:2504.07266}} (\bibinfo{year}{2025}).
\newblock


\bibitem[Yadollahi et~al\mbox{.}(2024)]%
        {10.1145/3610978.3638154}
\bibfield{author}{\bibinfo{person}{Elmira Yadollahi}, \bibinfo{person}{Marta Romeo}, \bibinfo{person}{Fethiye~Irmak Dogan}, \bibinfo{person}{Wafa Johal}, \bibinfo{person}{Maartje De~Graaf}, \bibinfo{person}{Shelly Levy-Tzedek}, {and} \bibinfo{person}{Iolanda Leite}.} \bibinfo{year}{2024}\natexlab{}.
\newblock \showarticletitle{Explainability for Human-Robot Collaboration}. In \bibinfo{booktitle}{\emph{Companion of the 2024 ACM/IEEE International Conference on Human-Robot Interaction}} (Boulder, CO, USA) \emph{(\bibinfo{series}{HRI '24})}. \bibinfo{publisher}{Association for Computing Machinery}, \bibinfo{address}{New York, NY, USA}, \bibinfo{pages}{1364–1366}.
\newblock
\showISBNx{9798400703232}
\href{https://doi.org/10.1145/3610978.3638154}{doi:\nolinkurl{10.1145/3610978.3638154}}


\end{thebibliography}

\end{document}